# Simple, Accurate, and Robust Nonparametric Blind Super-Resolution

Wen-Ze Shao, Michael Elad

Technion-Israel Institute of Technology, Department of Computer Science, Haifa, Israel
shaowenze1010@163.com, elad@cs.technion.ac.il

**Abstract.** This paper proposes a simple, accurate, and robust approach to single image blind super-resolution (SR). This task is formulated as a functional to be minimized with respect to both an intermediate super-resolved image and a *nonparametric* blur-kernel. The proposed method includes a convolution consistency constraint which uses a non-blind learning-based SR result to better guide the estimation process. Another key component is the bi-$\ell_0$-$\ell_2$-norm regularization placed on the super-resolved, sharp image and the blur-kernel, which is shown to be quite beneficial for accurate blur-kernel estimation. The numerical optimization is implemented by coupling the splitting augmented Lagrangian and the conjugate gradient. With the pre-estimated blur-kernel, the final SR image is reconstructed using a simple TV-based non-blind SR method. The new method is demonstrated to achieve better performance than Michaeli and Irani [2] in both terms of the kernel estimation accuracy and image SR quality.

## 1 Introduction

Since the seminal work by Freeman and Pasztor [3] and Baker and Kanade [4], single image super-resolution (SR) has drawn a considerable attention. A careful inspection of the literature in this area finds that existing approaches, either reconstruction-based or learning-based, focus on developing advanced image priors, however mostly ignoring the need to estimate the blur-kernel. Two recent comprehensive surveys on SR, covering work up to 2012 [5] and 2013 [33], testify that SR methods generally resort to the assumption of a known blur-kernel, both in the single image and the multi-image SR regimes. More specifically, in the context of multi-image SR, most methods assume a squared Gaussian kernel with a suitable standard deviation $\delta$, e.g., 3×3 with $\delta = 0.4$ [6], 5×5 with $\delta = 1$ [7], and so on. As for single image non-blind SR, we mention few commonly used options: bicubic low-pass filter (implemented by Matlab's default function *imresize*) [8-13, 21, 34, 35], 7×7 Gaussian kernel with $\delta = 1.6$ [13], 3×3 Gaussian kernel with $\delta = 0.55$ [14], and a simple pixel averaging kernel [15].

Interestingly, a related critical study on single image SR performance is presented in [1]. The authors have examined the effect of two components in single image SR, i.e., the choice of the image prior and the availability of an accurate blur model. Their conclusion, based on both the empirical and theoretical analysis, is that the influence of an accurate blur-kernel is significantly larger than that of an advanced image prior. Furthermore, [1] shows that "an accurate reconstruction constraint[1] combined with a simple gradient regularization achieves SR results almost as good as those of state-of-the-art algorithms with sophisticated image priors".

Only few works have addressed the estimation of an accurate blur model within the single image SR reconstruction process. Among few such contributions that attempt to estimate the kernel, a parametric model is usually assumed, and the Gaussian is a common choice, e.g., [16, 17, 36]. However, as the assumption does not coincide with the actual blur model, e.g., combination of out-of-focus and camera shake, we will naturally get low-quality SR results.

This paper focuses on the general single image *nonparametric* blind SR problem. The work reported in [18] is such an example, and actually it does present a nonparametric kernel estimation method for blind SR and blind deblurring in a unified framework. However, it is restricting its treatment to single-mode blur-kernels. In addition, [18] does not originate from a rigorous optimization principle, but rather builds on the detection and prediction of step edges as an important clue for the blur-kernel estimation. Another noteworthy and very relevant work is the one by Michaeli and Irani [2]. They exploit an inherent recurrence property of small natural image patches across different scales, and make use of the $MAP_k$-based estimation procedure [19] for recovering the kernel. Note that, the effectiveness of [2] largely relies on the found nearest neighbors to the query low-res patches in the input blurred, low-res image. We should also note that, in both [18] and [2] an $\ell_2$-norm-based kernel gradient regularization is imposed for promoting kernel smoothness.

Surprisingly, in spite of the similarity, it seems there exists a big gap between blind SR and blind image deblurring. The attention given to nonparametric blind SR is very small, while the counterpart blind deblurring problem is very popular and extensively treated. Indeed, a considerable headway has been made since Fergus *et al.*'s influential work [20] on camera shake removal. An extra down-sampling operator in the observation model is the only difference between the two tasks, as both are highly ill-posed problems, which admit possibly infinite solutions. A naturally raised hope is to find a unified and rigorous treatment for both problems, via exploiting appropriate common priors on the image and the blur-kernel.

Our contribution in this paper is the proposal of a simple, yet quite effective framework for general nonparametric blind SR, which aims to serve as an empirical answer towards fulfilling the above hope. Specifically, a new optimization functional is proposed for single image nonparametric blind SR. The blind deconvolution emerges naturally as a special case of our formulation. In the new approach, the first key component is harnessing a state-of-the-art non-blind dictionary-based SR method, generating a super-resolved but blurred image which is used later to constrain the blind SR.

The second component of the new functional is exploiting the bi-$\ell_0$-$\ell_2$-norm regularization, which was previously developed in [31] and imposed on the sharp image and the blur-kernel for blind motion deblurring[2]. We demonstrate that this unnatural prior along with a convolution consistency constraint, based on the super-resolved but blurred image, serve quite well for the task of accurate and robust nonparametric blind SR. This suggests that appropriate unnatural priors, especially on the images, are effective for both blind SR and blind deblurring. In fact, it has become a common belief in the blind deblurring community that [31, 22, 25, 26, 27] unnatural image priors are more essential than a natural one, be it a simple gradient-based or a complex learning-based prior.

---

[1] I.e., knowing the blur kernel.

We solve the new optimization functional in an alternatingly iterative manner, estimating the blur-kernel and the intermediate super-resolved, sharp image by coupling the splitting augmented Lagrangian (SAL) and the conjugate gradient (CG). With the pre-estimated blur-kernel, we generate the final high-res image using a simpler reconstruction-based non-blind SR method [38], regularized by the natural hyper-Laplacian image prior [31, 32, 37]. Comparing our results against the ones by [2] with both synthetic and realistic low-res images, our method is demonstrated to achieve quite comparative and even better performance in both terms of the blur-kernel estimation accuracy and image super-resolution quality.

The rest of the paper is organized as follows. Section 2 details the motivation and formulation of the proposed nonparametric blind SR approach, along with an illustrative example for a closer look at the new method. In Section 3, the numerical scheme with related implementation details for the optimization functional is presented. Section 4 provides the blind SR results by the proposed approach and [2], with both synthetic and realistic low-res images. Section 5 finally concludes the paper.

## 2 The Proposed Approach

In this section we formulate the proposed approach as a *maximum a posteriori* (MAP) based optimization functional. Let **o** be the low-res image of size $N_1 \times N_2$, and let **u** be the corresponding high-res image of size $sN_1 \times sN_2$, with $s > 1$ an up-sampling integer factor. The relation between **o** and **u** can be expressed in two ways:

$$\mathbf{o} = \mathbf{DKu} + \mathbf{n} \qquad (1)$$

$$\mathbf{o} = \mathbf{DUk} + \mathbf{n} \qquad (2)$$

where **U** and **K** are assumed to be the BCCB[3] convolution matrices corresponding to vectorized versions of the high-res image **u** and the blur-kernel **k**, and **D** represents a down-sampling matrix. In implementation, image boundaries are smoothed in order to prevent border artifacts. Our task is to estimate **u** and **k** given only the low-res image **o** and the up-sampling factor $s$.

In the non-blind SR setting, the work reported in [1] suggests that a simpler image gradient-based prior (e.g., the hyper-Laplacian image prior [32, 37]) can perform near-

---

[2] In [31] the bi-$\ell_0$-$\ell_2$-norm regularization is shown to achieve state-of-the-art kernel estimation performance. Due to this reason as well as the similarity between blind deblurring and blind SR, we extend the bi-$\ell_0$-$\ell_2$-norm regularization for the nonparametric blind SR problem.

[3] BCCB: block-circulant with circulant blocks.

ly as good as advanced learning-based SR models. In such a case, the restoration of **u** is obtained by

$$\hat{\mathbf{u}} = \arg\min_{\mathbf{u}} \ \lambda \| \mathbf{DKu} - \mathbf{o} \|_2^2 + \sum_j \rho((\nabla \mathbf{u})_j), \tag{3}$$

where $\rho$ is defined as $\rho(z) = |z|^\alpha$ with $0 \ll \alpha \leq 1$ leading to a sparseness-promoting prior, $\nabla = (\nabla_h; \nabla_v)$ with $\nabla_h, \nabla_v$ denoting the 1rst-order difference operators in the horizontal and vertical directions, respectively, and $\lambda$ is a positive trade-off parameter. In this paper, the fast non-blind SR method [38] based on the total variation prior (TV; $\alpha = 1$) is used for final SR image reconstruction. Nevertheless, the blind case is more challenging, and a new perspective is required to the choice of image and kernel priors for handling the nonparametric blind image SR.

### 2.1 Motivation and MAP Formulation

It is clear from (1) and (2) that the blur-kernel information is hidden in the observed low-res image. Intuitively, the accuracy of the blur-kernel estimation heavily relies on the quality of its counterpart high-res image that is reconstructed alongside with it. In blind deconvolution, it is generally agreed [19, 22] that commonly used natural image priors are likely to fail in recovering the true blur-kernel, as these priors prefer a blurred image over a sharp one. This applies not only to the simple $\ell_\alpha$-norm-based sparse prior ($0 \ll \alpha \leq 1$), but also the more complex learning-based Fields of Experts [23] as well as its extension [24], and so on. As a consequence, unnatural sparse image priors are more advocated recently in the blind deblurring literature [31, 22, 25, 26, 27].

Due to the close resemblance between blind image SR and the simpler blind deconvolution problem, and the fact that SR is more ill-posed, the same rationale is expected to hold for both problems, implying that we should use a "more extreme" prior for the high-res image. We note, however, that this refers to the first phase of blind image SR, i.e., the stage of blur-kernel estimation. Such an unnatural prior would lead to salient edges free of staircase artifacts which in turn are highly effective as core clues to blur-kernel estimation. It is natural that this would sacrifice some weak details in the high-res image, but as we validate hereafter, more precise and robust blur-kernel estimation can be achieved this way.

Prior to introducing our advocated image and kernel priors for the blind image SR task, we discuss another term to be incorporated into our MAP formulation. We assume the availability of an off-the-shelf fast learning-based SR method that is tuned to a simple and narrow bicubic blur. In this paper three candidate methods are considered, including: Neighborhood Embedding (NE) [21], Joint Sparse Coding (JSC) [10], and Anchored Neighbor Regression (ANR) [11]. Because the bicubic low-pass filter does not coincide with most realistic SR scenarios, such an algorithm generally generates a super-resolved but blurred image, denotes as $\tilde{\mathbf{u}}$. The relation between $\tilde{\mathbf{u}}$ and the unknown high-res image **u** can be roughly formulated as $\mathbf{Ku} \approx \tilde{\mathbf{u}}$. Therefore, we simply force a convolution consistency constraint to our MAP formulation, which results in an optimization problem of the form

$$\min_{\mathbf{u},\mathbf{k}} \ \lambda \| \mathbf{DKu} - \mathbf{o} \|_2^2 + \mathcal{R}_0(\mathbf{u},\mathbf{k}) + \eta \| \mathbf{Ku} - \tilde{\mathbf{u}} \|_2^2, \tag{4}$$

where $\eta$ is a positive trade-off tuning parameter, and $\mathcal{R}_0(\mathbf{u},\mathbf{k})$ is the image and kernel prior to be depicted in subsection 2.2. We set $\lambda = 0.01$, $\eta = 100$ for all the experiments in this paper. We emphasize that the convolution consistency constraint has greatly helped in decreasing unpleasant jagged artifacts in the intermediate super-resolved, sharp image $\mathbf{u}$, driving the overall minimization procedure to a better blur-kernel estimation.

## 2.2 Bi-$\ell_0$-$\ell_2$-Norm Regularization for Nonparametric Blind SR

The unnatural image priors that have been proven effective in blind deconvolution are those that approximate the $\ell_0$-norm in various ways [25, 26, 27, 22]. Instead of struggling with an approximation to the $\ell_0$-norm, in this paper, just like in [31], our strategy is to regularize the MAP expression by a direct bi-$\ell_0$-$\ell_2$-norm regularization, applied to both the image and the blur-kernel. Concretely, the regularization is defined as

$$\mathcal{R}_0(\mathbf{u},\mathbf{k}) = \alpha_\mathbf{u}(\|\nabla \mathbf{u}\|_0 + \tfrac{\beta_\mathbf{u}}{\alpha_\mathbf{u}}\|\nabla \mathbf{u}\|_2^2) + \alpha_\mathbf{k}(\|\mathbf{k}\|_0 + \tfrac{\beta_\mathbf{k}}{\alpha_\mathbf{k}}\|\mathbf{k}\|_2^2), \tag{5}$$

where $\alpha_\mathbf{u}, \beta_\mathbf{u}, \alpha_\mathbf{k}, \beta_\mathbf{k}$ are some positive parameters to be provided.

In Equation (5), the first two terms correspond to the $\ell_0$-$\ell_2$-norm-based image regularization. The underlying rationale is the desire to get a super-resolved, sharp image with salient edges from the original high-res image, which have governed the primary blurring effect, while also to force smoothness along prominent edges and inside homogenous regions. It is natural that such a sharp image is more reliable for recovering the true support of the desired blur-kernel than the ones with unpleasant staircase and jagged artifacts, requiring a kernel with a larger support to achieve the same amount of blurring effect. According to the parameter settings, a larger weight is placed on the $\ell_2$-norm of $\nabla \mathbf{u}$ than its $\ell_0$-norm, reflecting the importance of removing stair-case artifacts for smoothness in the kernel estimation process.

Similarly, the latter two terms in (5) correspond to the $\ell_0$-$\ell_2$-norm regularization for the blur-kernel. We note that the kernel regularization does not assume any parametric model, and hence it is applicable to diverse scenarios of blind SR. For scenarios such as motion and out-of focus blur, the rationale of the kernel regularization roots in the sparsity of those kernels as well as their smoothness. Compared against the $\ell_0$-$\ell_2$-norm image regularization, the $\ell_0$-$\ell_2$-norm kernel regularization plays a refining role in sparsitification of the blur-kernel, hence leading to an improved estimation precision. The $\ell_0$-norm part penalizes possible strong and moderate isolated components in the blur-kernel, and the $\ell_2$-norm part suppresses possible faint kernel noise, just as practiced recently in the context of blind motion deblurring in [26]. We should note that beyond the commonly used $\ell_2$-norm regularization, there are a few blind deblurring methods that use $\ell_1$-norm as well, e.g. [20, 25].

Now, we turn to discuss the choice of appropriate regularization parameters in Equation (5). Take the $\ell_0$-$\ell_2$-norm-based image regularization for example. If $\alpha_\mathbf{u}, \beta_\mathbf{u}$ are set too small throughout iterations, the regularization effect of sparsity promotion will be so minor that the estimated image would be too blurred, thus leading to poor quality estimated blur-kernels. On the contrary, if $\alpha_\mathbf{u}, \beta_\mathbf{u}$ are set too large, the intermediate sharp image will turn to too "cartooned", which generally has fairly less accurate edge structures accompanied by unpleasant staircase artifacts in the homogeneous regions, thus degrading the kernel estimation precision. To alleviate this problem, a continuation strategy is applied to the bi-$\ell_0$-$\ell_2$-norm regularization so as to achieve a compromise. Specifically, assume that current estimates of the sharp image and the kernel are $\mathbf{u}_i$ and $\mathbf{k}_i$. The next estimate, $\mathbf{u}_{i+1}$, $\mathbf{k}_{i+1}$, are obtained by solving a modified minimization problem of (4), i.e.,

$$(\mathbf{u}_{i+1}, \mathbf{k}_{i+1}) = \arg\min_{\mathbf{u},\mathbf{k}} \lambda \| \mathbf{DKu} - \mathbf{o} \|_2^2 + \mathcal{R}_1^i(\mathbf{u},\mathbf{k}) + \eta \| \mathbf{Ku} - \tilde{\mathbf{u}} \|_2^2, \tag{6}$$

where $\mathcal{R}_1^i(\mathbf{u},\mathbf{k})$ is given by

$$\mathcal{R}_1^i(\mathbf{u},\mathbf{k}) = c_\mathbf{u}^i \cdot \alpha_\mathbf{u} (\| \nabla \mathbf{u} \|_0 + \tfrac{\beta_\mathbf{u}}{\alpha_\mathbf{u}} \| \nabla \mathbf{u} \|_2^2) + c_\mathbf{k}^i \cdot \alpha_\mathbf{k} (\| \mathbf{k} \|_0 + \tfrac{\beta_\mathbf{k}}{\alpha_\mathbf{k}} \| \mathbf{k} \|_2^2), \tag{7}$$

where $c_\mathbf{u}$, $c_\mathbf{k}$ are the continuation factors, which are respectively set as $2/3$, $4/5$, and $c_\mathbf{u}^i$ denotes $c_\mathbf{u}$ to the power of $i$ [4]; as for the regularization parameters $\alpha_\mathbf{u}$, $\beta_\mathbf{u}$, $\alpha_\mathbf{k}$, $\beta_\mathbf{k}$, they are uniformly set as $\alpha_\mathbf{u} = 1$, $\beta_\mathbf{u} = 10$, $\alpha_\mathbf{k} = 0.2$, $\beta_\mathbf{k} = 1$ for all the experiments in this paper. With this continuation strategy, the regularization effect is diminishing as we iterate, which leads to more and more accurate salient edges in a progressive manner, and is shown quite beneficial for improving the blur-kernel estimation precision.

We will demonstrate hereafter that the proposed regularization (7) plays a vital role in achieving high estimation accuracy for the blur-kernel, and an $\ell_0$-norm-based image prior alone is not sufficient for serving this task.

### 2.3 A Closer Look at the Proposed Approach

To get a better insight for the proposed regularization on the sharp image and the blur-kernel, an illustrative example is provided in this subsection, relying on the numerical scheme to be presented in Section 3. Equation (7) is analyzed in a term-by-term way with three of its representative reduced versions studied, i.e.,

$$\mathcal{R}_2^i(\mathbf{u},\mathbf{k}) = c_\mathbf{u}^i \cdot \alpha_\mathbf{u} (\| \nabla \mathbf{u} \|_0) + c_\mathbf{k}^i \cdot \alpha_\mathbf{k} (\| \mathbf{k} \|_0 + \tfrac{\beta_\mathbf{k}}{\alpha_\mathbf{k}} \| \mathbf{k} \|_2^2), \tag{8}$$

$$\mathcal{R}_3^i(\mathbf{u},\mathbf{k}) = c_\mathbf{u}^i \cdot \alpha_\mathbf{u} (\| \nabla \mathbf{u} \|_0) + c_\mathbf{k}^i \cdot \beta_\mathbf{k} (\| \mathbf{k} \|_2^2), \tag{9}$$

$$\mathcal{R}_4^i(\mathbf{u},\mathbf{k}) = c_\mathbf{u}^i \cdot \alpha_\mathbf{u} (\| \nabla \mathbf{u} \|_0). \tag{10}$$

Naturally, several other reduced versions of Equation (7) can be tried as well; we select (8)-(10)[5] just for the convenience of presentation and illustration. With the given parameter values in subsection 2.2, we demonstrate that the success of Equation (7) depends on the involvement of all the parts in the regularization term. In addition, the superiority of the continuation strategy as explained above is validated. Actually, a si-milar analysis has been made in the context of blind motion deblurring [31], demonstrating well the effectiveness of the bi-$\ell_0$-$\ell_2$-norm regularization.

---

[4] The same meaning applies to $c_\mathbf{k}^i$.

[5] We should note that we have also selected a uniform set of parameter values for each of the formulations (8), (9) and (10), respectively, in order to optimize the obtained blind SR performance on a series of experiments. However, it was found that these alternative are still inferior to (7), just similar to the observation made in blind motion deblurring [31].

In **Fig. 1**, a low-res version of the benchmark high-res image Lena is provided, that is blurred by a 7×7 Gaussian kernel with $\delta = 1.5$ and down-sampled by a factor 2. We note that other blur-kernel types are tried in Section 4. Since we are blind to the kernel size, we just assume it to be 31×31. The SSD metric (Sum of Squared Difference)[19] is utilized to quantify the error between the estimated blur-kernel and its counterpart ground truth. For every regularization option, i.e., Equations (7)-(10), we test each of the three non-blind SR approaches, NE, JSC, ANR, for generating $\tilde{u}$. We also test the overall scheme without the continuation – this is denoted in the figure as 5-NE, 5-JSC, and 5-ANR.

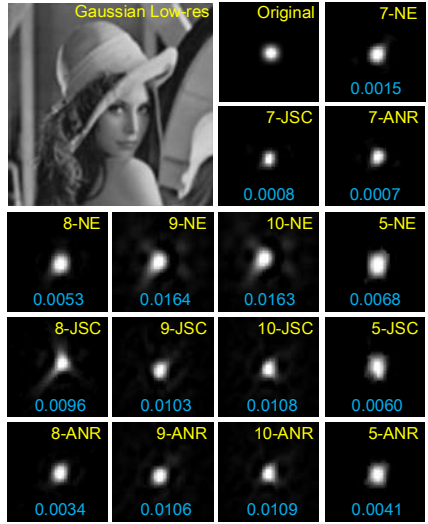

**Fig. 1.** An illustrative example of the bi-$\ell_0$-$\ell_2$-norm regularization for nonparametric blur-kernel estimation in single image blind SR. This figure shows the two times interpolated low-res image Lena (Nearest Neighbor), the ground truth blur-kernel, and the estimated ones using regularizations (7)-(10) with NE, JSC, ANR for generating the reference image $\tilde{u}$. The parts denoted by 5-NE/JSC/ANR correspond to the full scheme without continuation.

Clearly, the regularization by Equation (7) achieves the highest estimation accuracy compared to its degenerated versions. Take ANR for example: the SSD corresponding to 7-ANR is 0.0008, while those of 8-ANR, 9-ANR, 10-ANR and 5-ANR are 0.0034, 0.0106, 0.0109, and 0.0041, respectively. It is visually clear that the kernel of 5-ANR has a larger support than that of 7-ANR, validating the negative effect of the naive $\ell_0$-norm without continuation on the kernel estimation. Also, from the result of 10-ANR we deduce that the $\ell_0$-norm-based prior (with continuation) alone, is not sufficient. As incorporating other regularization terms into Equation (10), particularly the $\ell_0$-norm-based kernel prior and the $\ell_2$-norm-based image prior, higher estimation precision can be achieved because of the sparsification on the blur-kernel and the smoothness along dominant edges and inside homogenous regions of the image. Lastly, we note that it is crucial to incorporate the convolution consistency constraint based on an off-the-shelf non-blind SR method: when $\eta$ is set to 0 while other parameters in (6) are unaltered, the SSD of the estimated kernel increases to 0.0057.

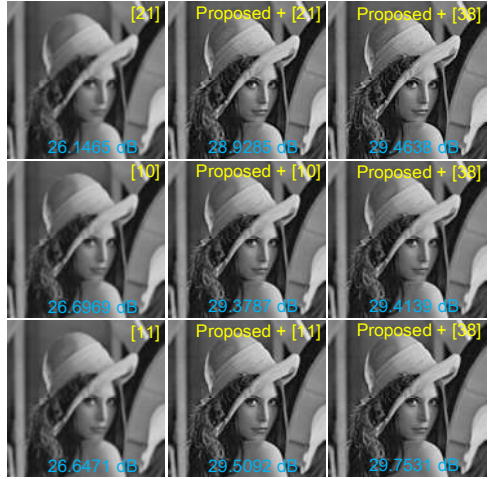

**Fig. 2.** Super-resolved images. First column: Non-blind results using the NE [21], JSC [10] and ANR [11] algorithms with the default bicubic blur-kernel. Second column: Blind results using [21, 10, 11] with blur-kernels estimated from the proposed method respectively based on 7-NE, 7-JSC, and 7-ANR. Third column: Blind results using the TV-based SR approach [38] with the estimated kernels.

In **Fig. 2**, super-resolved high-res images are estimated using learning-based non-blind SR algorithms [21, 10, 11], based on both the default bicubic low-pass filter and the kernels estimated by 7-NE, 7-JSC, 7-ANR shown in **Fig. 1**. It is clear that the super-resolved images shown in the second column of **Fig. 2** are of much better visual perception and higher PSNR (peak signal-to-noise ratio) than those shown in the first column. We note that ANR [11] (29.5092 dB) performs slightly better than JSC [10] (29.3787 dB) when fed with our estimated blur-kernels, and both approaches are superior to NE [21] (28.92858 dB), which accords with the experimental results in [11] that assume the known blur-kernels. It is also interesting to note that the TV-based SR method [38] (i.e., the third column in **Fig. 2**), along with our estimated blur-kernels, achieves better performance than all the candidate non-blind SR methods [21, 10, 11], among which the proposed 7-ANR+[38] ranks the best (29.7531 dB). Recall the claim in [1] that an accurate reconstruction constraint plus a simpler $\ell_\alpha$-norm-based sparse image prior is almost as good as state-of-the-art approaches with sophisticated image priors. This aligns well with the results shown here. In Section 4, we utilize [38] for the final non-blind SR image reconstruction.

## 3   Numerical Algorithm

We now discuss the numerical aspect of minimizing the non-convex and non-smooth functional (6). Because of the involved $\ell_0$-norm, the optimization task is generally NP-hard. We do not attempt to provide a rigorous theoretical analysis on the existence of a global minimizer of (6) or make a claim regarding the convergence of the proposed numerical scheme. We do note, however, that there are few encouraging attempts that shed some theoretical light on problems of related structure to the one posed here (see [28, 29]). Nevertheless, considering the blind nature of our problem, the focus here is on a practical numerical algorithm.

### 3.1 Alternating Minimization

We formulate the blur-kernel estimation in (6) as an alternating $\ell_0$-$\ell_2$-regularized least-squares problem with respect to **u** and **k**. Given the blur-kernel $\mathbf{k}_i$, the super-resolved, sharp image **u** is estimated via

$$\mathbf{u}_{i+1} = \arg\min_{\mathbf{u}} \ \alpha_{\mathbf{u}} \|\nabla \mathbf{u}\|_0 + \beta_{\mathbf{u}} \|\nabla \mathbf{u}\|_2^2 + \tfrac{\lambda}{c_{\mathbf{u}}^i} \|\mathbf{D}\mathbf{K}_i \mathbf{u} - \mathbf{o}\|_2^2 + \tfrac{\eta}{c_{\mathbf{u}}^i} \|\mathbf{K}_i \mathbf{u} - \tilde{\mathbf{u}}\|_2^2. \quad (11)$$

Turning to estimating the blur-kernel $\mathbf{k}_{i+1}$ given the image $\mathbf{u}_{i+1}$, our empirical experimentation suggests that this task is better performed when implemented in the image derivative domain. Thus, $\mathbf{k}_{i+1}$ is estimated via

$$\begin{aligned}\mathbf{k}_{i+1} = \arg\min_{\mathbf{k}} \ & \alpha_{\mathbf{k}} \|\mathbf{k}\|_0 + \beta_{\mathbf{k}} \|\mathbf{k}\|_2^2 \\ & + \sum_{d \in \Lambda} \left\{ \tfrac{\lambda}{c_{\mathbf{k}}^i} \|\mathbf{D}(\mathbf{U}_{i+1})_d \mathbf{k} - \mathbf{o}_d\|_2^2 + \tfrac{\eta}{c_{\mathbf{k}}^i} \|(\mathbf{U}_{i+1})_d \mathbf{k} - \tilde{\mathbf{u}}_d\|_2^2 \right\}, \end{aligned} \quad (12)$$

subject to the constraint set $\mathcal{C} = \{\mathbf{k} \geq 0, \|\mathbf{k}\|_1 = 1\}$, since a blur-kernel should be non-negative as well as normalized. In Equation (12), $(\mathbf{U}_{i+1})_d$ represents the convolution matrix corresponding to the image gradient $(\mathbf{u}_{i+1})_d = \nabla_d \mathbf{u}_{i+1}$, $\mathbf{o}_d = \nabla_d \mathbf{o}$, $\tilde{\mathbf{u}}_d = \nabla_d \tilde{\mathbf{u}}$.

Both (11) and (12) can be solved in the same manner as in [30, 31] based on the splitting augmented Lagrangian (SAL) approach. The augmented Lagrangian penalty parameters for (11) and (12) are set as $\gamma_{\mathbf{u}} = 100$, $\gamma_{\mathbf{k}} = 1 \times 10^6$, respectively. Note that, due to the involved down-sampling operator, we use the CG method to calculate each iterative estimate of **u** or **k**. In the CG, the error tolerance and the maximum number of iterations are set respectively as $1e^{-5}$ and 15.

### 3.2 Multi-scale Implementation

In order to make the proposed approach adaptive to large-scale blur-kernels as well as to reduce the risk of getting stuck in poor local minima when solving (11) and (12), a multi-scale strategy is exploited. For clarity, the pseudo-code of multi-scale implementation of the proposed approach is summarized as **Algorithm 1**

In each scale, the low-res image **o** and the super-resolved, but blurred image $\tilde{\mathbf{u}}$ are down-sampled two times accordingly as inputs to (11) and (12). In the finest scale the inputs are the original **o** and $\tilde{\mathbf{u}}$ themselves. The initial image for each scale is simply set as the down-sampled version of $\tilde{\mathbf{u}}$, and the initial blur-kernel is set as the bicubic up-sampled kernel produced in the coarser scale (in the coarsest scale it is simply set as a Dirac pulse).

**Algorithm 1. Alternating minimization for nonparametric blind SR**

1: **Input.** Images $\mathbf{o}$, $\tilde{\mathbf{u}}$, down-sampled images $\mathbf{o}_s$ and $\tilde{\mathbf{u}}_s$ in coarser scales $s < 4$, and $\mathbf{o}_4 = \mathbf{o}$, $\tilde{\mathbf{u}}_4 = \tilde{\mathbf{u}}$.
2: **Initialization.** $s = 1$, $i = 0$, $\mathbf{u}_0 = \tilde{\mathbf{u}}_1$, $\mathbf{k}_0 =$ Dirac pulse.
3: **While** $s \leq 4$, do
    4: **While** $i < 10$, do
        5: • Solve (11) for $\mathbf{u}_{i+1}$ with 10 iterations of SAL;
        6: • Solve (12) for $\mathbf{k}_{i+1}$ with 10 iterations of SAL;
    7: **End**
    8: • $i = 0$;
    9: • Set $\mathbf{k}_0$ by upsampling $\mathbf{k}_{10}$ with projection onto $\mathcal{C}$ for the $(s+1)^{\text{th}}$ scale;
    10: • Set $\mathbf{u}_0$ by $\tilde{\mathbf{u}}_s$ for the $(s+1)^{\text{th}}$ scale;
11: **End**
13: **Output :** $\hat{\mathbf{k}}$.
14: **Non-blind SR:** Super-resolve the final high-res image $\hat{\mathbf{u}}$ using the TV-based SR method [38] with the estimated kernel $\hat{\mathbf{k}}$.

## 4 Experimental Results

This section validates the benefit of the proposed approach using both synthetic and realistic low-res images[6]. The non-blind SR method ANR [11] is chosen for conducting the blur-kernel estimation in all the experiments. We make comparisons between our approach and the recent state-of-the-art nonparametric blind SR method reported in [2]. It is noted that the estimated blur-kernels corresponding to [2] were prepared by Tomer Michaeli who is the first author of [2]. Due to this comparison, and the fact that the work in [2] loses its stability for large kernels[7], we restrict the size of the kernel to 19×19. In spite of this limitation, we will try both 19×19 and 31×31 as the input kernel sizes to our proposed approach, just to verify its robustness against the kernel size.

The first group of experiments is conducted using ten test images from the Berkeley Segmentation Dataset, as shown in **Fig. 3**. Each one is blurred respectively by a 7×7, 11×11, and 19×19 Gaussian kernel with $\delta = 2.5$, 3 times down-sampled, and degraded by a white Gaussian noise with noise level equal to 1. Both the image PSNR and the kernel SSD are used for quantitative comparison between our method and [2]. **Table 1** presents the kernel SSD (scaled by 1/100), and **Table 2** provides the PSNR scores of correspondingly super-resolved images by the non-blind TV-based SR approach [38] with the kernels estimated in **Table 1**. From the experimental results, our method in both kernel sizes, i.e., 19×19, 31×31, achieves better performance than [2] in both the kernel SSD and the image PSNR. We also see that, as opposed to the sensitivity of the method in [2], our proposed method is robust with respect to the input kernel size.

---

[6] Experiments reported in this paper are performed with MATLAB v7.0 on a computer with an Intel i7-4600M CPU (2.90 GHz) and 8 GB memory.
[7] In [2] blur-kernels are typically solved with size 9×9, 11×11 or 13×13 for various blind SR problems.

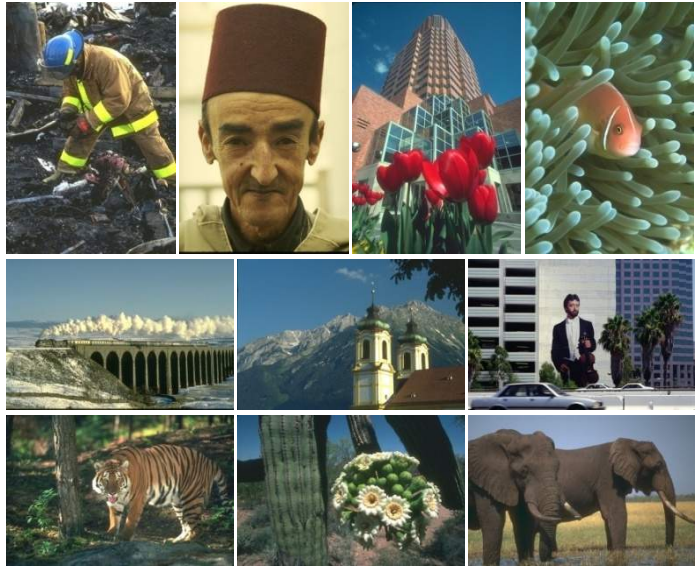

**Fig. 3.** Test images from the Berkeley Segmentation Dataset used for quantitative evaluation of each nonparametric blind SR method. Left to right, top to bottom: (*a*)-(*j*).

**Table 1.** SSD of the blur-kernels estimated by [2] and our method. "Ours.1" corresponds to our method with an input kernel size of 19×19, and "Ours.2" corresponds to the size 31×31.

| True size | ×10⁻² | *a* | *b* | *c* | *d* | *e* | *f* | *g* | *h* | *i* | *j* | Mean |
|---|---|---|---|---|---|---|---|---|---|---|---|---|
| 7×7 | [2] | 0.97 | 0.47 | 1.06 | 0.77 | 1.28 | 0.83 | 1.13 | 1.08 | 1.46 | 0.26 | **0.93** |
| | Ours.1 | 0.22 | 0.13 | 0.22 | 0.20 | 0.22 | 0.19 | 0.22 | 0.21 | 0.17 | 0.15 | **0.19** |
| | Ours.2 | 0.23 | 0.14 | 0.25 | 0.15 | 0.27 | 0.17 | 0.23 | 0.22 | 0.18 | 0.13 | **0.20** |
| 11×11 | [2] | 0.29 | 0.18 | 0.42 | 0.32 | 0.70 | 0.56 | 0.87 | 0.41 | 0.62 | 0.10 | **0.45** |
| | Ours.1 | 0.13 | 0.02 | 0.10 | 0.07 | 0.08 | 0.07 | 0.11 | 0.15 | 0.05 | 0.04 | **0.08** |
| | Ours.2 | 0.11 | 0.04 | 0.09 | 0.08 | 0.07 | 0.05 | 0.07 | 0.13 | 0.05 | 0.03 | **0.09** |
| 19×19 | [2] | 0.22 | 0.15 | 0.38 | 0.26 | 0.63 | 0.30 | 0.83 | 0.35 | 0.55 | 0.09 | **0.38** |
| | Ours.1 | 0.11 | 0.03 | 0.09 | 0.07 | 0.08 | 0.07 | 0.14 | 0.16 | 0.07 | 0.04 | **0.09** |
| | Ours.2 | 0.09 | 0.05 | 0.11 | 0.08 | 0.07 | 0.06 | 0.13 | 0.14 | 0.08 | 0.03 | **0.08** |

**Fig. 4** shows SR results for a synthetically blurred image, with a severe motion blur. This example demonstrates well the robustness of the proposed approach to the kernel type, while either the non-blind ANR [11] or the blind method [2] completely fails in achieving acceptable SR performance. **Fig. 5** and **Fig. 6** present blind SR results on two realistic images (downloaded from the Internet). The image in **Fig. 5** is somewhat a mixture of motion and Gaussian blur. We see that both our method and [2] produce reasonable SR results, while ours is of relatively higher quality; the faces in the super-resolved image with our estimated kernels can be better recognized to a great degree. As for **Fig. 6**, our method also produces a visually more pleasant SR image, while the jagged and ringing artifacts can be clearly observed in the SR image corresponding to

[2], which produces an unreasonable blur-kernel. Please see the SR images on a computer screen for better perception.

**Table 2.** PSNR of correspondingly super-resolved images by the non-blind TV-based SR approach [38] with the estimated kernels in **Table 1**

| True size | dB | a | b | c | d | e | f | g | h | i | j | Mean |
|---|---|---|---|---|---|---|---|---|---|---|---|---|
| 7×7 | [2] | 21.0 | 25.8 | 22.7 | 23.0 | 21.1 | 25.6 | 21.2 | 22.9 | 22.4 | 27.7 | **23.3** |
|  | Ours.1 | 24.9 | 27.9 | 24.0 | 30.6 | 23.4 | 27.7 | 22.6 | 25.0 | 25.8 | 29.1 | **26.1** |
|  | Ours.2 | 24.9 | 27.8 | 24.2 | 30.5 | 22.7 | 27.7 | 22.1 | 25.1 | 25.7 | 28.8 | **26.0** |
| 11×11 | [2] | 21.7 | 25.9 | 22.7 | 23.5 | 20.6 | 25.5 | 21.1 | 23.3 | 22.5 | 27.7 | **23.5** |
|  | Ours.1 | 24.6 | 27.9 | 24.3 | 30.1 | 23.7 | 27.5 | 22.8 | 25.0 | 25.6 | 29.0 | **26.1** |
|  | Ours.2 | 24.6 | 27.7 | 24.3 | 30.0 | 23.7 | 27.5 | 22.6 | 25.0 | 25.6 | 28.9 | **26.0** |
| 19×19 | [2] | 21.7 | 26.1 | 22.6 | 23.8 | 20.7 | 25.2 | 21.1 | 23.4 | 22.6 | 27.8 | **23.5** |
|  | Ours.1 | 24.6 | 27.9 | 24.4 | 30.2 | 23.8 | 27.5 | 22.8 | 24.9 | 25.6 | 28.9 | **26.1** |
|  | Ours.2 | 24.6 | 27.6 | 24.4 | 30.2 | 23.8 | 27.5 | 22.7 | 25.0 | 25.6 | 28.9 | **26.0** |

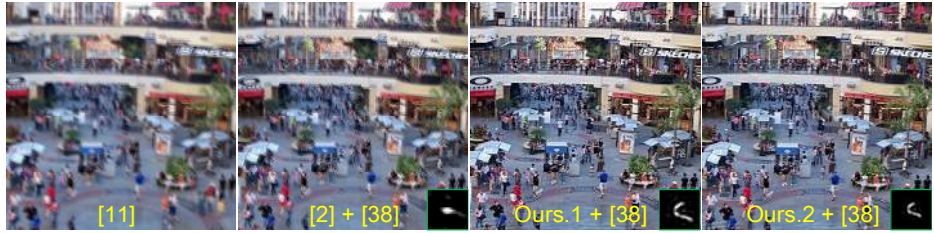

**Fig. 4.** SR with synthetic low-res Hollywood (×2). Left to right: Non-blind ANR [11]; [2]+[38] (size 19×19); Ours.1+[38] (size 19×19); Ours.2+ [38] (size 31×31).

## 5    Conclusions and Discussions

This paper presents a new method for nonparametric blind SR, formulated as an optimization functional regularized by a bi-$\ell_0$-$\ell_2$-norm of both the image and blur-kernel. Compared with the state-of-the-art method reported in [2], the proposed approach is shown to achieve quite comparative and even better performance, in both terms of the blur-kernel estimation accuracy and the super-resolved image quality.

An elegant benefit of the new method is its relevance for both blind deblurring and blind SR reconstruction, treating both problems in a unified way. Indeed, the bi-$\ell_0$-$\ell_2$-norm regularization, primarily deployed in [31] for blind motion deblurring, proves its effectiveness here as well, and hence serves as the bridge between the two works and the two problems. The work can be also viewed as a complement to that of [1] in providing empirical support to the following two claims: (*i*) blind SR prefers appropriate unnatural image priors for accurate blur-kernel estimation; and (*ii*) a natural prior, no matter be it simple (e.g., $\ell_\alpha$-norm-based sparse prior [32]) or advanced (e.g., Fields of Experts [23]), are more appropriate for non-blind SR reconstruction.

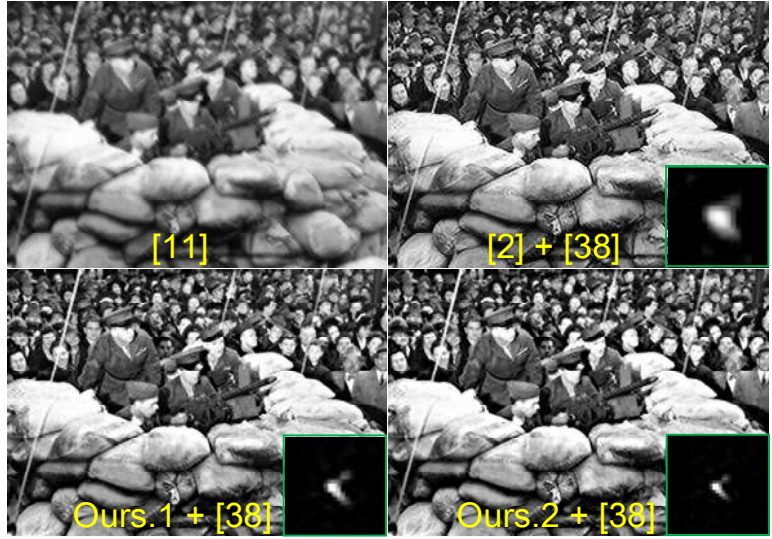

**Fig. 5.** SR with low-res Crowd (×2). Top left: Non-blind ANR [11]; Top right: [2]+[38] (size 19×19); Bottom left: Ours.1+[38] (size 19×19); Bottom right: Ours.2+[38] (size 31×31).

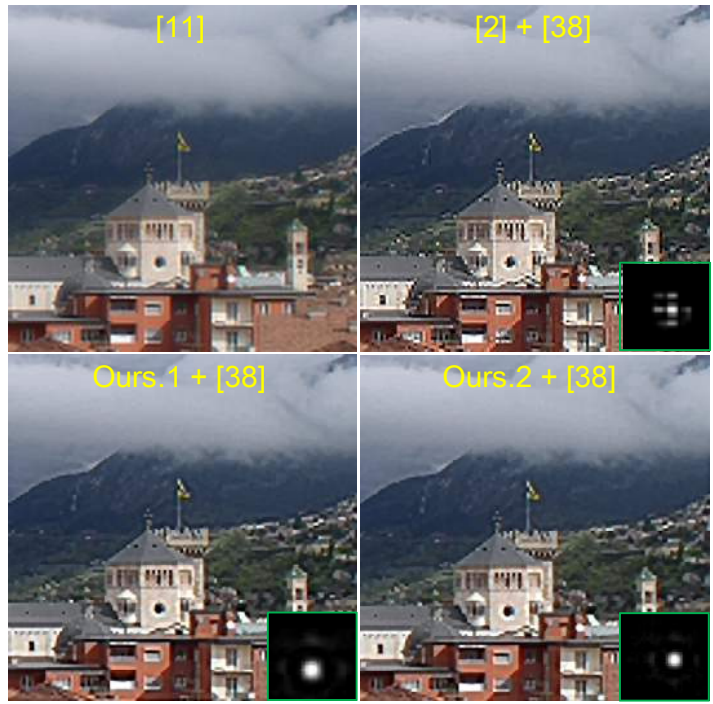

**Fig. 6.** SR with low-res Building (×4). Top left: Non-blind ANR [11]; Top right: [2]+[38] (size 19×19); Bottom left: Ours.1+[38] (size 19×19); Bottom right: Ours.2+[38] (size 31×31).

## Acknowledgements

We would like to thank Dr. Tomer Michaeli for his kind help in running the blind SR method [2], enabling the reported comparison between the proposed approach and [2]. The first author is thankful to Professor Zhi-Hui Wei, Professor Yi-Zhong Ma, Dr. Min Wu, and Mr. Ya-Tao Zhang for their kind support in the past years. This research was supported by the European Research Council under EU's 7th Framework Program, ERC Grant agreement no. 320649, the Google Faculty Research Award, the Intel Collaborative Research Institute for Computational Intelligence, and the Natural Science Foundation (NSF) of China (61402239), the NSF of Government of Jiangsu Province (BK20130868), and the NSF for Jiangsu Institutions (13KJB510022), and the NSF of NUPT (NY212014, NY213134).